\documentclass{article}

\usepackage[final]{neurips_ai4mat_2023}

\usepackage{algorithm}
\usepackage[noend]{algpseudocode}
\usepackage[utf8]{inputenc} 
\usepackage[T1]{fontenc}    
\usepackage{hyperref}       
\usepackage{url}            
\usepackage{booktabs}       
\usepackage{amsfonts}       
\usepackage{nicefrac}       
\usepackage{microtype}      
\usepackage{xcolor}         
\usepackage{graphicx} 
\usepackage{doi}
\usepackage{xcolor}
\usepackage{subcaption}
\usepackage{array}
\usepackage{adjustbox} 
\usepackage{amsmath}
\usepackage{pifont}
\usepackage[flushleft]{threeparttable}
\usepackage{arydshln}
\newcommand{\cmark}{\ding{51}}%
\newcommand{\xmark}{\ding{55}}%

\title{Gotta be \emph{SAFE}: A New Framework for Molecular Design}

\author{%
  Emmanuel~Noutahi \\
  Valence Labs\\
  Montr\'eal, QC, Canada \\
  \texttt{emmanuel@valencelabs.com} \\
  \And
  Cristian~Gabellini \\
  Valence Labs\\
  Montr\'eal, QC, Canada \\
  \texttt{cristian@valencelabs.com} \\
  \And
  Michael~Craig \\
  Valence Labs\\
  Montr\'eal, QC, Canada \\
  \texttt{michael@valencelabs.com} \\
  \And
  Jonathan~S.C~Lim \\
  Mila \& Valence Labs\\
  Montr\'eal, QC, Canada \\
  \texttt{jonathan.lim@u.nus.edu} \\
  \And
  Prudencio~Tossou \\
  Valence Labs\\
  Montr\'eal, QC, Canada \\
  \texttt{prudencio@valencelabs.com} \\
}

\begin{document}

\hypersetup{
pdftitle={Gotta be SAFE: A new Framework for Molecular Design},
pdfsubject={cs.LG},
pdfauthor={Emmanuel~Noutahi},
pdfkeywords={SMILES strings, scaffold, SAFE, linker, generative models, transformers}
}
\maketitle

\begin{abstract}
Traditional molecular string representations, such as SMILES, often pose challenges for AI-driven molecular design due to their non-sequential depiction of molecular substructures. To address this issue, we introduce \textbf{S}equential \textbf{A}ttachment-based \textbf{F}ragment \textbf{E}mbedding (SAFE), a novel line notation for chemical structures. SAFE reimagines SMILES strings as an unordered sequence of interconnected fragment blocks while maintaining compatibility with existing SMILES parsers. It streamlines complex generative tasks, including scaffold decoration, fragment linking, polymer generation, and scaffold hopping, while facilitating autoregressive generation for fragment-constrained design, thereby eliminating the need for intricate decoding or graph-based models. We demonstrate the effectiveness of SAFE~\footnote{Code, data and model available at https://github.com/datamol-io/safe/} by training an 87-million-parameter GPT2-like model on a dataset containing 1.1 billion SAFE representations. Through targeted experimentation, we show that our SAFE-GPT model exhibits versatile and robust optimization performance. SAFE opens up new avenues for the rapid exploration of chemical space under various constraints, promising breakthroughs in AI-driven molecular design. 
\end{abstract}

\section{Introduction}
\label{sec:intro}
Molecular design, which consist of constructing molecules with desired characteristics, is a critical task in computational drug discovery. It often necessitates the preservation of certain scaffolds or core chemical substructures, which serve as the backbone for the design process, The motivation for preserving these groups and constraints typically stems from their crucial role in the molecule's biological activity. Nevertheless, incorporating such constraints can be challenging when relying on conventional molecular string representations like the Simplified Molecular Input Line Entry System (SMILES).

Although SMILES has played a crucial role in chemistry and drug discovery, it is unable to provide a contiguous representation of molecular substructures. This limitation hinders tasks like adding structures to a molecule's scaffold and connecting fragments, limiting its usefulness in improving potential drug candidates, particularly during lead optimization efforts. Addressing these challenges requires the development of an enhanced line notation for molecules, one that can preserve the integrity of molecular scaffolds and fragments while offering flexibility for \textit{de novo} molecular design.

\begin{figure}[!tb]
\centering
 \includegraphics[width=0.85\columnwidth]{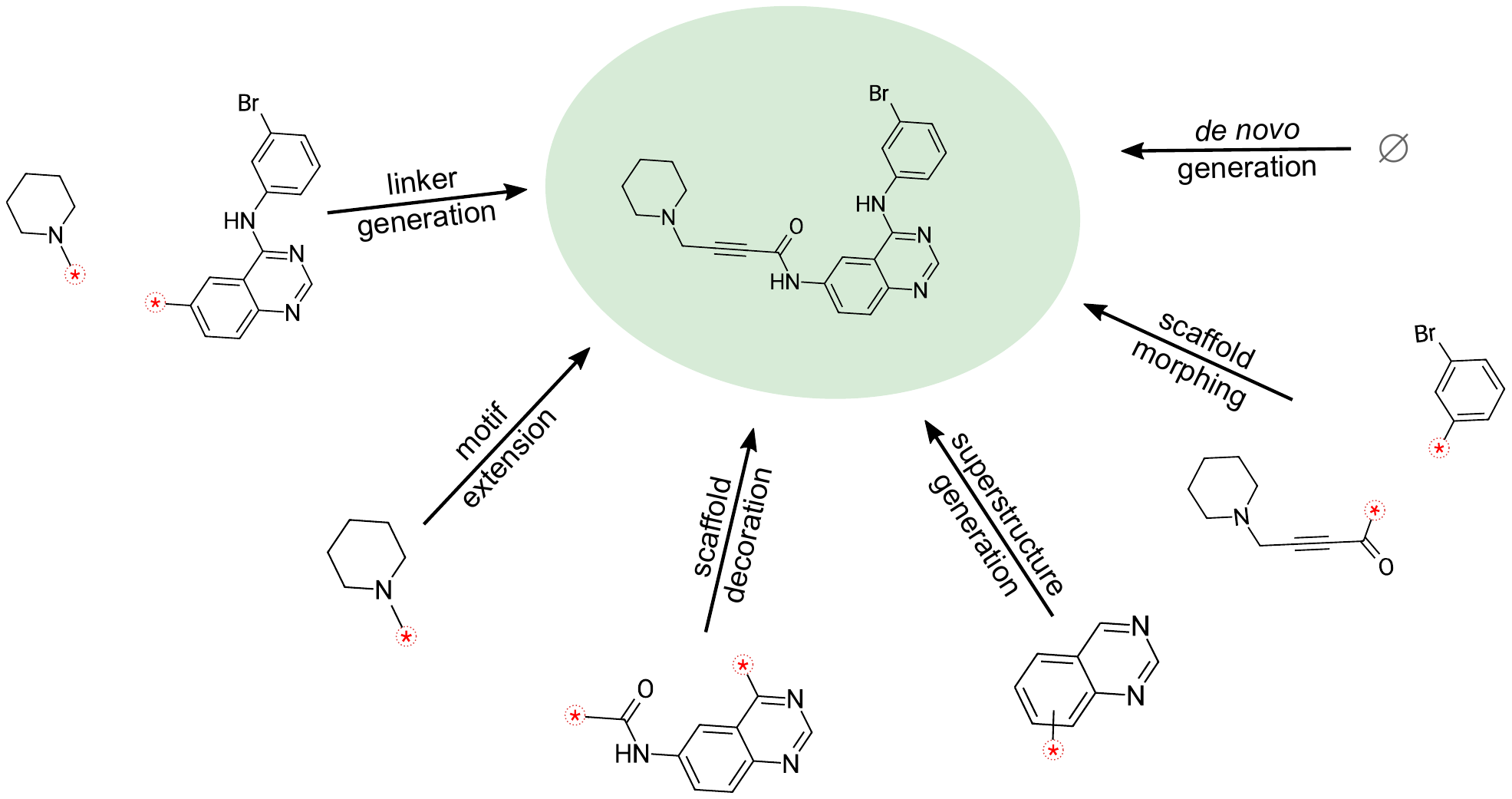}
	\caption{\small Molecular design tasks that can be performed easily with SAFE}
	\label{fig:mol-design}
\end{figure}

To this end, we introduce Sequential Attachment-based Fragment Embedding (SAFE), a novel line notation for molecules. In contrast to existing methods, SAFE represents molecules as an unordered sequence of fragment blocks. This re-imagines molecular design tasks, transforming them into simpler sequence completion problems. Moreover, SAFE facilitates autoregressive generation, effectively bypassing the need for intricate decoding schemes or graph-based models (see Figure~\ref{fig:mol-design} and \autoref{tab:generative-capabilities}). Importantly, despite these novel features, SAFE strings are backward compatible with SMILES parsers, promising an easy integration into existing workflows. Our contributions can be summarized as follow:
\begin{itemize}
\item We introduce SAFE, a novel molecular representation compatible with SMILES that represents molecules as a sequence of interconnected fragments.
\item We introduce SAFE-GPT, an 87.3-million-parameter GPT-like generative model, pretrained on a dataset of 1.1 billion SAFE strings that can be used for diverse downstream tasks. This model is shown to be effective in various molecule generation tasks, capitalizing on SAFE's unique characteristics.
\item We propose a new benchmark inspired by real-world drug discovery challenges to assess pure generative models' performance in tasks such as scaffold decoration, linker design, and motif extension.
\end{itemize}

\section{Related Works}
\label{sec:litterature}
\textbf{Molecular line notation representations:} The Simplified Molecular-Input Line-Entry System (SMILES) \citep{weininger1988smiles} is the most widely adopted molecular line notation in chemoinformatics for its simplicity, compactness, and human readability. In contrast to the International Chemical Identifier (InCHI) that provides global and unique identifier to molecules, SMILES are more suitable for molecular generation tasks. However, SMILES lack robustness to minor changes and struggle with ensuring the validity and integrity of fragments in deep learning-based molecular design. They also underperform in molecular search and substructure matching tasks. To overcome these challenges, alternative notations like Self-Referencing Embedded Strings (SELFIES) \citep{krenn2020self, krenn2022selfies} have been developed. SELFIES address the robustness and validity issues in deep generative modeling through a recursive approach, surparssing notations like DeepSmiles \citep{DeepSmiles} and GenSMILES \citep{genSmilesBHADWAL2023110429}, but come at the cost of simplicity, interpretability and compactness. None of these notations consistently uphold the integrity of scaffolds and fragments essential for several molecular generation tasks. A recent innovation, Group SELFIES\citep{GroupSELFIES2023}, builds on standard SELFIES by introducing functional and chemical group tokens, to improve compactness and chemical motif representation for molecular generative tasks. Yet, neither Group SELFIES nor other line notations facilitate deep generative fragment-based molecule design without extensive, task-specific engineering of training processes and molecule generation steps \citep{guo2023link,fialkova2021libinvent,Langevin2020,liao2023sc2mol}, bespoke model architectures \citep{arus2020smiles}, or goal-directed optimization frameworks. In \autoref{tab:generative-capabilities}, we contrast the generative capabilities of various molecular line notations, including SAFE.

\textbf{Deep generative design:} To contextualize our work within the domain of deep generative design we refer interested readers to comprehensive reviews provided in \citep{DavidL2020MolReprReview,Bilodeau2022Review,du2022molgensurvey}. Herein, we briefly describe sequence-based and graph-based deep generative models. 
Sequence-based methods, originally focused on character-by-character SMILES generation \citep{gomez2018automatic}. This approach provided considerable versatility but faced challenges when dealing with fragment-based constraints. Nevertheless, recent advancements have attempted to address this limitation by separately generating scaffolds and side chains \citep{liao2023sc2mol}, introducing transformations derived from matched molecular pairs analysis \citep{he2022transformer}, and employing conditional generation \citep{yang2021transformer, bagal2021molgpt}. 
In the realm of graph-based methods, our work shares similarities with \citep{pmlr-v80-jin18a,jin2020multi,maziarz2021learning}, which uses motifs for molecular graphs but encounter difficulties when extending design to scaffold-based generation, linker-design and generating molecules with unseen building blocks. In particular, these methods, while capable of assembling motifs in a tree-like structure, have difficulties creating novel cyclic structures not seen during training.




\textbf{Constrained molecular design:} Notable contributions have emerged in the recent literature on constrained molecular design. \cite{li2018multi} introduced a conditional graph generative model that excels in producing valid molecules while offering the flexibility needed for multi-objective optimization. MolGPT \citep{bagal2021molgpt}, which uses a transformer-decoder architecture for the generation of drug-like molecules, has demonstrated the capacity to conditionally control diverse molecular properties and scaffold designs, highlighting its efficacy in crafting molecules tailored to specific requirements. Furthermore, Multi-Constraint Molecular Generation (MCMG) \citep{wang2021multi}, combining conditional transformers, knowledge distillation, and reinforcement learning, has shown the capability to satisfy multiple constraints during the process of molecular generation.

\textbf{Scaffold-conditioned generation:} Under hard scaffold constraints, \cite{Lim_2020} proposed a graph-based model explicitly trained on scaffold and molecule pairs. Under soft scaffold constraints, \cite{li2018learning} have considered the scaffold as part of the input, but their approach does not guarantee its presence in the generated molecules. \cite{arus2020smiles}  used an iterative conditional training procedure to perform scaffold decoration with an LSTM trained on SMILES. Their work was extended in \citep{fialkova2021libinvent}, where a reaction-driven approach for scaffold decoration was proposed. Finally, \cite{Langevin2020} proposed a sampling algorithm that can adapt any SMILES-based auto-regressive model to work with scaffolds. However, being trained on SMILES, their models can neither guarantee validity of generated molecules nor the presence of the input scaffold constraint.

\begin{table}[!hb]
\small
\caption{Pure generative capabilities of various molecular representations. In the assessment of the inherent generative capabilities of each molecular representation, we employ a marking system: \cmark~signifies intrinsic competence, \textbf{?}~indicates the need for additional and intentional engineering, and \xmark~suggests unverified capabilities.\\}
\label{tab:generative-capabilities}
\centering
\begin{tabular}{m{2.5cm}llm{1.5cm}lm{1.5cm}ll}
  \toprule
  \textbf{Task} & \textbf{SAFE} & \textbf{SMILES} & \textbf{Deep/Gen SMILES} & \textbf{SELFIES} & \textbf{Group SELFIES} & \textbf{InChi} & \textbf{GRAPHS} \\
  \midrule
  De novo design & \cmark &  \cmark & \cmark & \cmark & \cmark & \textbf{?} & \cmark \\
  Linker design & \cmark &  \textbf{?} & \xmark & \xmark & \textbf{?} & \xmark & \textbf{?} \\
  Motif extension & \cmark & \textbf{?} & \xmark & \textbf{?} & \textbf{?} & \xmark & \cmark \\
  Scaffold decoration & \cmark &  \textbf{?} & \xmark & \xmark & \textbf{?} & \xmark & \cmark \\
  Scaffold morphing & \cmark &  \xmark & \xmark & \xmark & \textbf{?} & \xmark & \textbf{?} \\
  Super structure & \cmark &  \xmark & \xmark & \xmark & \textbf{?} & \xmark & \cmark \\
  \bottomrule
\end{tabular}
\end{table}

\section{SAFE algorithm}
\label{sec:safe}

\begin{figure}[!htb]
	\centering
\includegraphics[width=0.8\columnwidth]{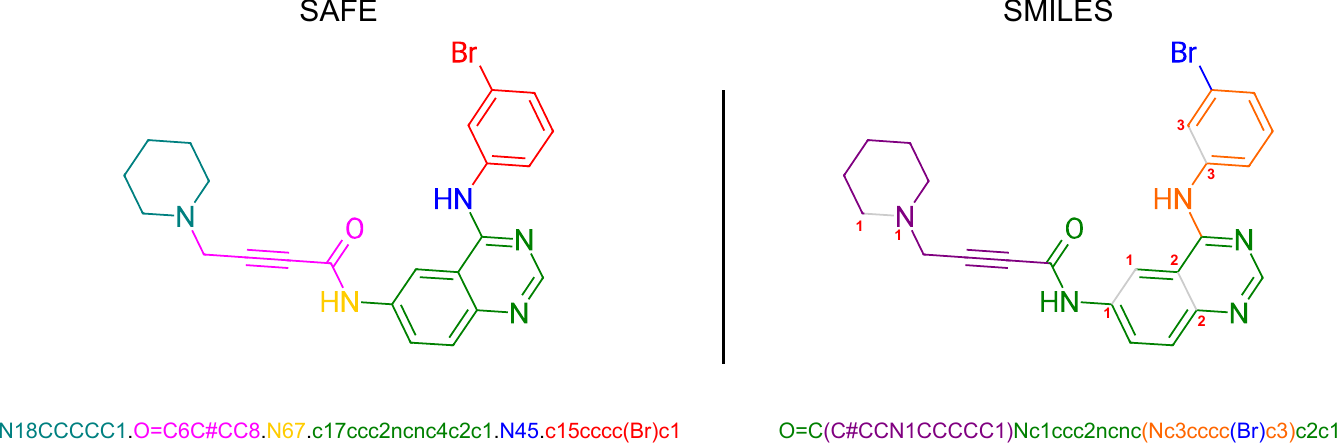}
	\caption{Example of a molecule as a SAFE and SMILES representation. The colored fragments and their corresponding placement in each string show how the ordering of the fragments in the SAFE representation are more easily readable and interpretable than the comparable SMILES string.}\label{fig:safe-vs-smiles}
\end{figure}

In SMILES, ring structures are marked by using digits to identify the opening and closing ring atom, thus denoting a virtual connection between the corresponding atoms. This rule also contributes to the surjectivity of SMILES representation where multiple different SMILES correspond to the same molecular graph. 
SAFE (Sequence Attachment-based Fragment Embedding) leverages this rule to discover alternative SMILES strings that enforce an order of SMILES characters in which all SMILES tokens belonging to the same molecular fragment are consistently arranged consecutively (see Figure~\ref{fig:safe-vs-smiles}). As such, SAFE is a molecular line notation that reimagines SMILES as a collection of connected fragments and remains a valid SMILES representation. Furthermore, the arrangement of fragments within a SAFE string has no impact on the underlying molecular graph, ensuring that common data augmentation techniques for generative models, such as randomization, remain applicable.

\subsection{Constructing A SAFE string}
The detailed process to convert from SMILES to SAFE is illustrated by Algorithm~\ref{algo:safe-construction} and Figure~\ref{fig:safe-build}.

\begin{algorithm}[!htb]
\caption{Conversion of SMILES to SAFE Representation}
\label{algo:safe-construction}
\small
\begin{algorithmic}[1] 
\Procedure {ToSAFE}{$molecule$}
\State $ring\_digits \gets$ extract all unique ring digits from $molecule$ 
\State $fragments \gets$ fragment $molecule$ on  specified bonds \Comment We use BRICS bonds here
\State Sort fragments in $fragments$ by size in descending order
\State $fragments\_str \gets \{\}$
\For {each $frag$ in $fragments$}
\State Add smiles of $frag$ to $fragments\_str$
\EndFor
\State $safe\_str \gets$ join all elements in $fragments\_str$ with "."
\State $attach\_pos \gets$ extract all attachment points from $safe\_str$

\State $i \gets$ $\max\big(ring\_digits\big)$ + 1 \Comment Find the next possible ring digits
\For {each $attach$ in $attach\_pos$}
\State Replace $attach$ in $safe\_str$ with $i$
\State Increment $i$ by 1
\EndFor
\State \Return $safe\_str$
\EndProcedure
\end{algorithmic}
\end{algorithm}

It starts by extracting all unique ring digits from the associated molecule and fragmenting it on a desired set of bonds. Our implementation utilizes the BRICS algorithm (\cite{degen2008art}), though other bond-splitting algorithms, such as Hussain-Rea~\citep{hussain2010computationally}, RECAP~\citep{lewell1998recap}, or custom patterns, are equally valid. These substructures may represent synthetically accessible building blocks that are common in drug-like compounds. The extracted fragments are sorted by size and concatenated, using a dot character  (\textbf{"."}) to mark new fragments in the representation, while preserving their corresponding attachment points. To construct the final SAFE string, we iterate over the numbered attachment points and replace them by novel ring digits to simulate fragment linking. These new ring digits create virtual connections between fragments resulting in a set of linked fragments, as indicated by the dot character. It's worth noting that, similar to canonicalization in SMILES that yields a unique representation from multiple valid forms, we can achieve a similar outcome by enforcing a decoding order not only on SMILES characters within fragments but also on fragment orders within the final SAFE string.

\subsection{SAFE facilitates fragment-based design}
\label{sec:safe-and-fragment-design}

The inherent sequential block structure of SAFE presents a distinctive advantage for fragment-based design tasks. Traditionally, such endeavors primarily relied on graph-based generative models. However, with a generative model trained on SAFE strings, fragment-based design becomes remarkably straightforward (refer to Figure~\ref{fig:mol-design}).

Among those, we found the following particularly suitable for SAFE:

\begin{itemize}
    \item \textbf{De novo generation:} which consists of sampling a new sequence from the learned token distribution. It's as straightforward with SAFE as with established SMILES-based auto-regressive models used in molecular generation.

    \item \textbf{Scaffold decoration and motif extension:} which can be framed as sequence completion and new tokens prediction to create novel fragments using SAFE. Starting with an initial sequence corresponding to a scaffold or motif, and marked attachment points for completion, SAFE simplifies this compared to other notations.

    \item \textbf{Linker design and scaffold morphing:} that can also be approached as sequence completion task.  Since the order of fragments in a SAFE string doesn't affect the underlying molecular graph, the fragments to be linked can be provided as the initial sequence for a generative model to predict likely tokens for the missing linker.
    
    \item \textbf{Superstructure generation:}  in this setting, the goal is to generate new molecules while adhering to a specified substructure constraint. In the SAFE framework, we achieve this by first generating random attachment points on the substructure to create new scaffolds, followed by scaffold decoration.
\end{itemize}

\section{Experiments}
\label{sec:experiments}
To evaluate the utility of our new molecular line notation, we developed a generative model using a decoder-only transformer architecture. Our aim is to showcase the model's ability, trained on SAFE strings, to generate valid and diverse molecules in \textit{de novo} scenarios. Additionally, we seek to evaluate its effectiveness in practical, real-world scenarios where tasks like scaffold decoration, scaffold morphing, linker design and goal-directed generation are required.

\subsection{SAFE-GPT: SAFE generative model}

\textbf{Dataset}: We began by constructing a vast chemical dataset comprising over 1 billion unlabeled molecules for pre-training purposes. This dataset was carefully constructed by combining molecules from the ZINC and UniChem libraries~\citep{irwin2005zinc,chambers2013unichem}, resulting in a diverse collection of 1.1 billion SMILES strings. Our dataset spans various molecule types, encompassing drug-like compounds, peptides, multi-fragment molecules, polymers, reagents and non-small molecules,  ensuring the wide applicability of our generative model. It stands as the largest and most diverse dataset designed specifically for deep generative molecular design. To convert SMILES strings into SAFE strings, we utilized a combination of BRICS decomposition and a graph partitioning method (Louvain community detection), when BRICS bonds where not available. Molecules that couldn't undergo successful fragmentation were excluded from our dataset. For our experiments we do not use randomization of fragment positions or SMILES ordering due to the already large dataset.

\textbf{Tokenizer}: We trained a BPE tokenizer on the full dataset. As a pre-tokenization step for the inputs, we applied a common regular expression for SMILES syntax~\citep{schwaller2019molecular}. This process yielded a vocabulary of 1180 tokens, including all special tokens (\textit{EOS, BOS, UNK, MASK, PAD}).

\textbf{Model architecture}: Our SAFE Generative model (SAFE-GPT) is a 87.3M parameters GPT2-like transformer. It comprises 12 layers, each with 12 attention heads per layer, and a hidden state size of 768. All other model parameters adhere to the default settings of GPT-2, as outlined in Hugging Face.

\textbf{Model training}: The SAFE Model (SAFE-GPT) was trained using cross-entropy with the next token prediction as training objective. We use  the AdamW optimizer ($\beta_1 = 0.9$ and $\beta_2 = 0.999$)~\citep{kingma2014adam},
a linear learning rate scheduler with 10000 warmup steps and an initial $lr=1e-4$. We set the batch size to 100 per GPU and used 
2 steps of gradient accumulation and gradient checkpointing. The model was trained on 4 Nvidia A100 GPUs, for a maximum of 1000000 steps (7 days). 

\textbf{SAFE and Group SELFIES GPT-20M models on MOSES dataset}: Additionally, we trained a smaller 20M-parameters (6 layers, 8 attention heads per layer, and a hidden state size of 768) version of SAFE-GPT (SAFE-GPT-20M), and a Group SELFIES version with the same architecture (GSELFIES-GPT-20M) on the MOSES dataset \citep{2020mosesbenchmarking} for comparative analysis. These models were trained for 10 epochs, using similar loss functions, optimizer configurations as SAFE-GPT but with an initial $lr=5e-4$. We followed the Group SELFIES original implementation for tokenization. For a detailed comparison between the performance of SAFE-GPT-20M and GSELFIES-GPT-20M, refer to~\autoref{appendix:safe-groupselfies-comparison}.
 
\subsection{De novo generation results}

In \textit{de novo} design, our objective is to generate entirely novel compounds with desirable profiles. Assessing a model's ability to generate valuable compounds in such a setting, even without an optimization objective is crucial, as some models may encounter problems generating valid or sufficiently diverse and novel compounds. We used classical metrics like molecule validity, uniqueness, and internal diversity \citep{2020mosesbenchmarking,Huang2021tdc} to assess these qualities. \textit{Validity} measures the percentage of chemically valid structures according to the RDKit's parser, \textit{Uniqueness} is the fraction of non-duplicate molecules, and \textit{Diversity} assesses the internal diversity of generated molecules using the average pairwise Tanimoto distance (ECFP4 representation).

\begin{table}[!htb]
\centering
\begin{threeparttable}
\caption{\textbf{Molecule generation results on 10K samples}. The large pretrained SAFE-GPT model performs similarly to models trained on the MOSES dataset while producing more diverse molecules.\\
}
\label{table:1}
\begin{tabular}{lllll} 
\toprule
 \textbf{Model} & \textbf{Repr.} & \textbf{Valid@10K $\uparrow$} & \textbf{Unique@10k $\uparrow$}   & \textbf{Diversity $\uparrow$} \\ 
 \midrule
 \textbf{SAFE-GPT\tnote{*}} & SAFE & 0.984 & \textbf{1} & 0.878 \\ 
 \hdashline
 \textbf{SAFE-GPT-20M} & SAFE & \textbf{1} & 0.999 & 0.864 \\
 GSELFIES-GPT-20M & Group SELFIES & \textbf{1} & 0.999 & \textbf{0.887} \\
 GSELFIES-VAE & Group SELFIES & \textbf{1} & 0.999 & 0.859 \\
 GMT-SELFIES & SELFIES & \textbf{1} &  \textbf{1} & 0.870 \\
 SELFIES-VAE & SELFIES & \textbf{1} & 0.999 & 0.858 \\
 CharRNN & SMILES & 0.975 &  0.999 & 0.856 \\ 
 VAE & SMILES & 0.977 & 0.998 & 0.856 \\ 
 LatentGAN & SMILES & 0.897 &  0.997 & 0.857 \\ 
 LigGPT & SMILES & 0.900 &  0.999 & 0.871 \\
 JT-VAE & GRAPH & \textbf{1} & 0.999 & 0.855 \\ 

 \bottomrule
\end{tabular}%
\begin{tablenotes}
  \small
  \item[*] SAFE-GPT uses a different training dataset that includes non drug-like and challenging molecules.
\end{tablenotes}
\end{threeparttable}
\end{table}

Table~\ref{table:1} showcases a comparison of SAFE-GPT with various generative models across 10,000 samples. Despite being trained on a dataset encompassing challenging molecules, SAFE-GPT still demonstrates impressive performance in validity, uniqueness, and diversity. Remarkably, it surpasses other models in uniqueness and diversity, although it has a marginally lower validity score. To determine if this is linked to the complexities in interpreting fragment connectivity, represented by digit pairs—a common challenge also observed in SMILES-based models—we trained a smaller version, SAFE-GPT-20M, on the MOSES dataset, as well as an alternative model with same architecture that uses Group SELFIES representation (GSELFIES-GPT-20M). The $100\%$ validity observed for SAFE-GPT-20M suggests that SAFE-GPT's slightly reduced validity is largely due to its diverse and challenging training dataset. Compared to SAFE-GPT models, GSELFIES-GPT-20M appears to generate more diverse molecules. However, a closer examination of its outputs (refer to \autoref{appendix:safe-groupselfies-comparison}) reveals a tendency to create large, unstable rings in otherwise "valid" chemical graphs, leading to very low druglikeness and synthetic accessibility.

In \autoref{fig:sample-denovo}, we show a subset of randomly selected molecules generated with SAFE-GPT. This visual representation offers readers an intuitive sense of the quality and reasonableness of the generated molecules. Furthermore, in Figure~\ref{fig:sample-prop-dist}, we show the distribution of selected molecular properties for the 10,000 generated molecules.

\subsection{Performance on fragment-constrained generation}

\textit{De novo} compound generation is only one approach for advancing a drug discovery program. In fact, in many real-world scenarios, generative design involves modifying existing molecules in user-defined ways rather than creating entirely new compounds. This is especially true in later stages of drug discovery, such as hit-to-lead or lead optimization, where well-established structure-activity relationships (SAR) are already in place. Therefore, we examined SAFE's intended capabilities for performing fragment-constrained generative design tasks such as scaffold decoration, scaffold morphing, linker generation, motif extension, and superstructure generation (see \autoref{sec:safe-and-fragment-design}). To facilitate this evaluation, we designed a benchmark that involved working with scaffolds and fragments from 10 existing drugs. Further details about the benchmark design can be found in~\autoref{appendix:frag-benchmark} in the Appendix. Our focus on SAFE-GPT is due to its unique capability to perform these tasks without substantial modifications in the representation, training, or sampling process. In fact attempts at performing those tasks with the Group SELFIES model (GSELFIES-GPT-20M) mostly resulted in a failure to maintain the fragment constraints. Although we were able to perform the superstructure tasks, the generated samples by the Group SELFIES model exhibit very low uniqueness ($6\%$) and low internal diversity ($0.43$).

Table~\ref{table:fragment-design-results} presents averaged validity, diversity, and uniqueness scores for 1000 molecules sampled in each fragment-constrained design task using SAFE-GPT across all drugs. It displays the average Tanimoto distance between the generated molecules to the original drug molecules, along with the average SA score (Synthetic Accessibility Score) \citep{ertl2009estimation}, which we used the RDKit library \citep{greg_landrum_2023_10099869_rdkit} to generate. We observe that SAFE-GPT maintains full validity for all sampled molecules under constraints, while achieving high internal diversity and novelty compared to the original drugs. Moreover, generated molecules exhibit a low SA score, indicating their ease of synthesis. For a visual inspection of sample molecules from each task using Maribavir as the starting molecule, please refer to Table~\ref{table:example-sample-fragment-design} (\autoref{appendix:frag-benchmark}).

\begin{table}[!htb]
\small 
\centering
\caption{Performance on fragment-constrained generative design tasks on 1000 molecules sampled\newline}
\label{table:fragment-design-results}
\begin{tabular}{@{}llllll@{}} 
\toprule
 \textbf{Task} & \textbf{Validity $\uparrow$} & \textbf{Diversity $\uparrow$}   & \textbf{Uniqueness $\uparrow$}  & \textbf{Distance $\uparrow$} & \textbf{SA score $\downarrow$} \\ 
 \midrule
Linker design     &       1.000±0.000 &
                    0.641±0.099 &
                    0.887±0.191 &
                    0.712±0.097 &
                    3.864±0.928 \\
Motif extension    &        1.000±0.000 &
                    0.681±0.089 &
                    0.923±0.179 &
                    0.772±0.101 &
                    3.750±0.651 \\
Scaffold decoration   &       1.000±0.000 &
                    0.571±0.113 &
                    0.851±0.162 &
                    0.643±0.137 &
                    4.017±0.889 \\
Scaffold morphing    &       1.000±0.000 &
                    0.608±0.096 &
                    0.717±0.219 &
                    0.688±0.113 &
                    3.604±0.910\\
Superstructure  &    1.000±0.000 &
                    0.715±0.059 &
                    0.929±0.106 &
                    0.812±0.063 &
                    3.868±0.919\\

 \bottomrule
\end{tabular}%
\end{table}

\subsection{Goal-directed generative capabilities}
\label{goal-directed-physchem}
To effectively apply generative approaches in live drug discovery projects, it is essential to incorporate goal-directed generation, guiding generation of novel molecules towards specific properties. Therefore, we follow established methodologies~\citep{lim2020scaffold,seo2023molecular} to assess the model's ability for goal-directed generation using simple molecular properties. More precisely, we optimize toward specific values for key molecular properties, including Topological Polar Surface Area (TPSA), Molecular Weight (MW), Calculated LogP (CLOGP), and Quantitative Estimation of Drug-likeness (QED). To achieve this, we use Proximal Policy Optimization (PPO)~\citep{schulman2017proximal} with Adaptive KL Penalty to train a policy for generating molecular samples with the targeted property value. A total of 50 steps was performed with a learning rate of 1e-5 (AdamW optimizer) and a batch size of 100. The reward objective used for this optimization was defined as follows: 

\[\text{reward}(mol) = \frac{1}{1 + \alpha \cdot \left| \text{prop}(mol) - \text{target} \right|}\]

where \textit{prop(mol)} represents the calculated molecular property value for a given sample, \textit{target} signifies the desired target value, and $\alpha$ is set to 0.5.

\begin{figure}[!htb]
\centering
\includegraphics[width=1\columnwidth]{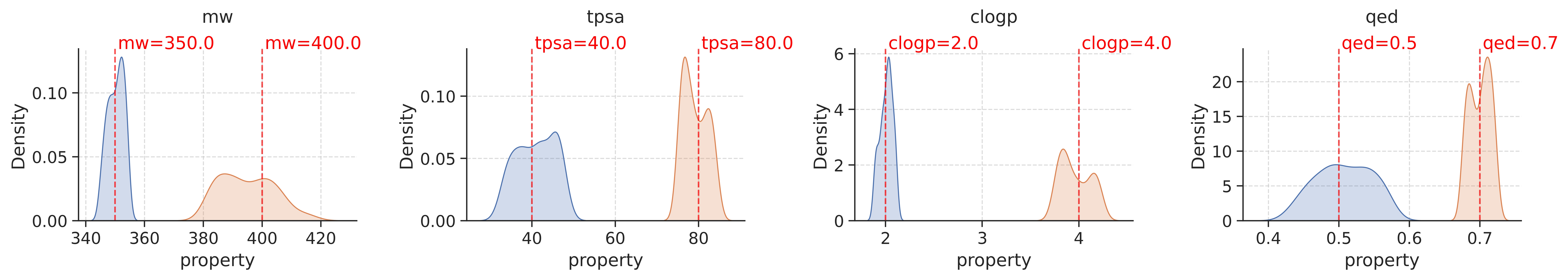}
    \caption{Property distributions of generated molecules, grouped by molecular properties, after goal-conditioned optimization using PPO. The red line in each plot shows the target value the agent was optimized towards.}
    \label{fig:fig-optim}
\end{figure}

With the methodology described above, we fine-tuned agents for two target values on each molecular property and evaluated their performance by generating 500 samples from each of them. Notably, all generated samples were valid and unique. The property distribution of these samples is visually presented in Figure~\ref{fig:fig-optim}, where the red line within each plot represents the target value of the molecular property that the agent was optimized towards, and the blue and orange histograms representing the distribution of samples from different agents with distinct goals. The results depicted in Figure~\ref{fig:fig-optim} demonstrate that the property distribution of the generated molecules, achieved through goal-conditioned optimization using PPO, is notably centered around the respective target values. This outcome indicates the success of our optimization process in aligning the generated molecules distribution with the desired property targets.

\subsection{Scaffold-Constrained optimization of CNS penetration of EGFR inhibitors}

In this section, we introduce a novel and challenging optimization task aimed at improving the Central Nervous System (CNS) penetration of EGFR Tyrosine Kinase Inhibitors. This optimization task specifically addresses the challenge of CNS metastases in non-small cell lung cancer, a significant concern in cancer treatment~\citep{ahluwalia2018epidermal}. Our objective involves optimizing the CNS-MPO score, a comprehensive metric assessing physico-chemical properties associated with CNS penetration~\citep{wager2016central}. The CNS-MPO score ranges from 0 to 6, with higher scores indicating better desirability, and a score above 4 typically suffices.
We introduce additional constraints to our optimization task, requiring that all generated molecules feature a scaffold that has demonstrated activity against EGFR (see ~\autoref{fig:cns-penetration}). For an in-depth exploration of this topic, please consult \autoref{cns-optimization-appendix} in the Appendix.

\begin{figure}[!htb]
\centering
\includegraphics[width=1\columnwidth]{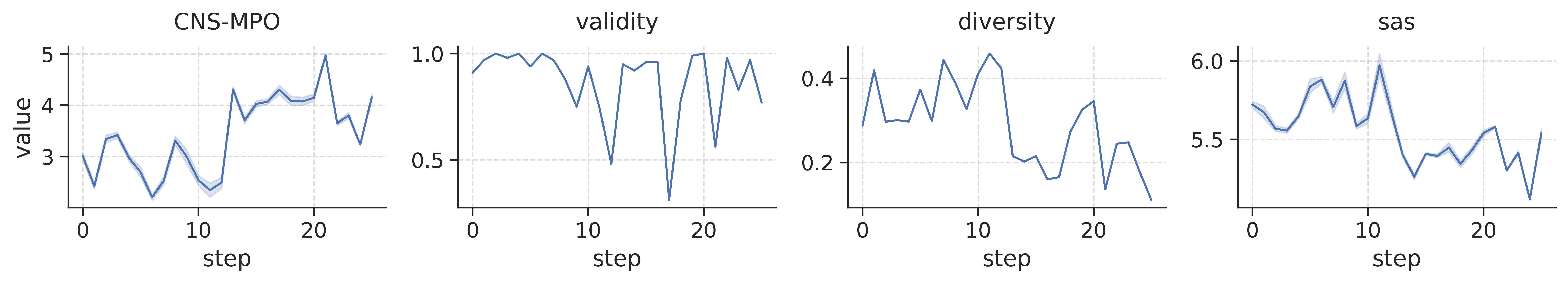}
    \caption{Distribution of CNS-MPO rewards and generative metrics score (validity, internal diversity and SA score) throughout the 25 optimization steps when sampling 100 molecules from the RL agent.}
    \label{fig:fig-optim-cns-mpo}
\end{figure}

We directly optimize the CNS-MPO score using PPO for 25 steps, and the same training parameters outlined in ~\autoref{goal-directed-physchem}.

Figure~\ref{fig:fig-optim-cns-mpo} illustrates the reward distribution obtained by sampling 100 molecules at each optimization iteration. Our findings demonstrate that scaffold-constrained optimization, even when facing challenging metrics, can be efficiently executed with SAFE-GPT using a straightforward optimization algorithm like PPO. As the CNS-MPO policy refines, we observe an expected reduction in the diversity of sampled candidates, while overall validity remains robust. Intriguingly, there's a slight decline in the SA score across iterations, suggesting the presence of synthetically favorable yet optimal compounds within the solution space.
  

\section{Discussion}
\label{sec:discussion}
This work introduces SAFE, a novel molecular representation that enhances versatility and expressive power in molecular design while retaining compatibility with SMILES parsers. SAFE represents molecules as sequences of interconnected fragments, offering a new paradigm in molecular description. It emerges as a promising alternative to existing molecular line notations, addressing their limitations by striking a balance between simplicity and robustness, thus making it suitable for a wide range of applications.

We also present SAFE-GPT, a pioneering generative model with 87.3 million parameters, trained on 1.1 billion diverse SAFE strings. The model's effectiveness in various generative and optimization tasks highlights SAFE's unique attributes. Although we observed slightly lower molecule validity in SAFE-GPT, this can be mostly attributed to the complexity and diversity of its training set. We posit that a better sampling algorithm, potentially enforcing phrasal constraints \citep{post2018fast} around digit tokens, could address this issue.

The potential for fine-tuning SAFE-GPT on specialized chemical spaces opens avenues for enhancing its utility in targeted tasks. While this work focuses on a benchmark set of 10 drugs for fragment-constrained generation, we plan to extend this to a broader range of drugs, providing a comprehensive evaluation of the model's capabilities in various molecular generation scenarios. In future works, we aim to explore SAFE's performance in multi-property optimization (MPO) scenarios, including the integration of a prediction head into the SAFE-GPT architecture for simultaneous molecular generation and property prediction. Ultimately, we seek to efficiently scale SAFE-GPT to larger models and datasets, laying the groundwork for a new generation of foundational models in drug discovery.

Our work brings significant advancements in molecular representation and generative modeling. We believe that these innovations will continue to drive progress in drug discovery, materials science, and other fields where molecular design plays a pivotal role.

\clearpage
\bibliographystyle{unsrtnat}
\bibliography{references} 

\clearpage
\appendix
\section{Supplementary Material}
\label{sec:appendix}
\subsection{Additional figures}

\begin{figure}[!htb]
	\centering
\includegraphics[width=0.9\columnwidth]{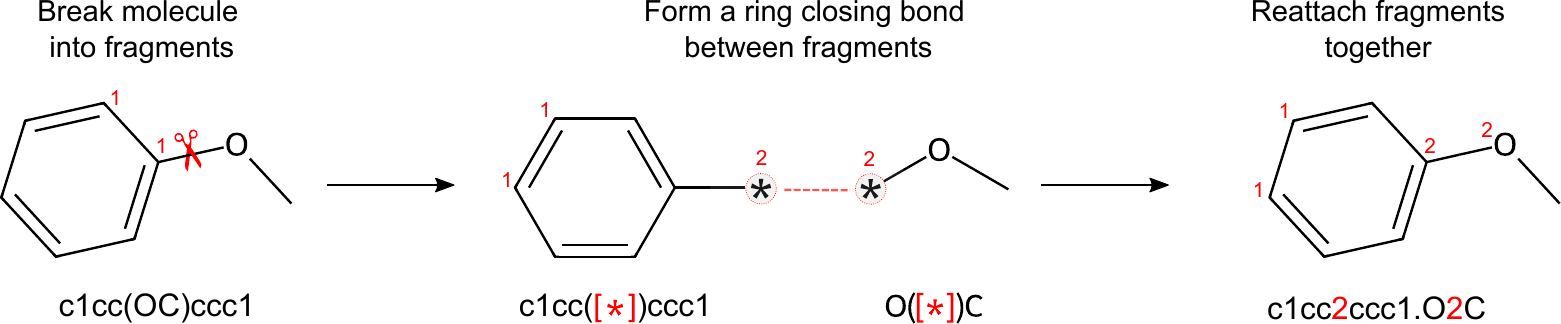}
	\caption{Example encoding of a SMILES string into a SAFE representation. The left panel shows the breaking a bond by the BRICS algorithm. The middle panel shows the addition of attachment points and the ring closing bond connecting the two fragments. The right panel shows the reattached fragments and the final SAFE representation.}
	\label{fig:safe-build}
\end{figure}

\begin{figure}[!htb]
    \centering
    \includegraphics[width=0.9\columnwidth]{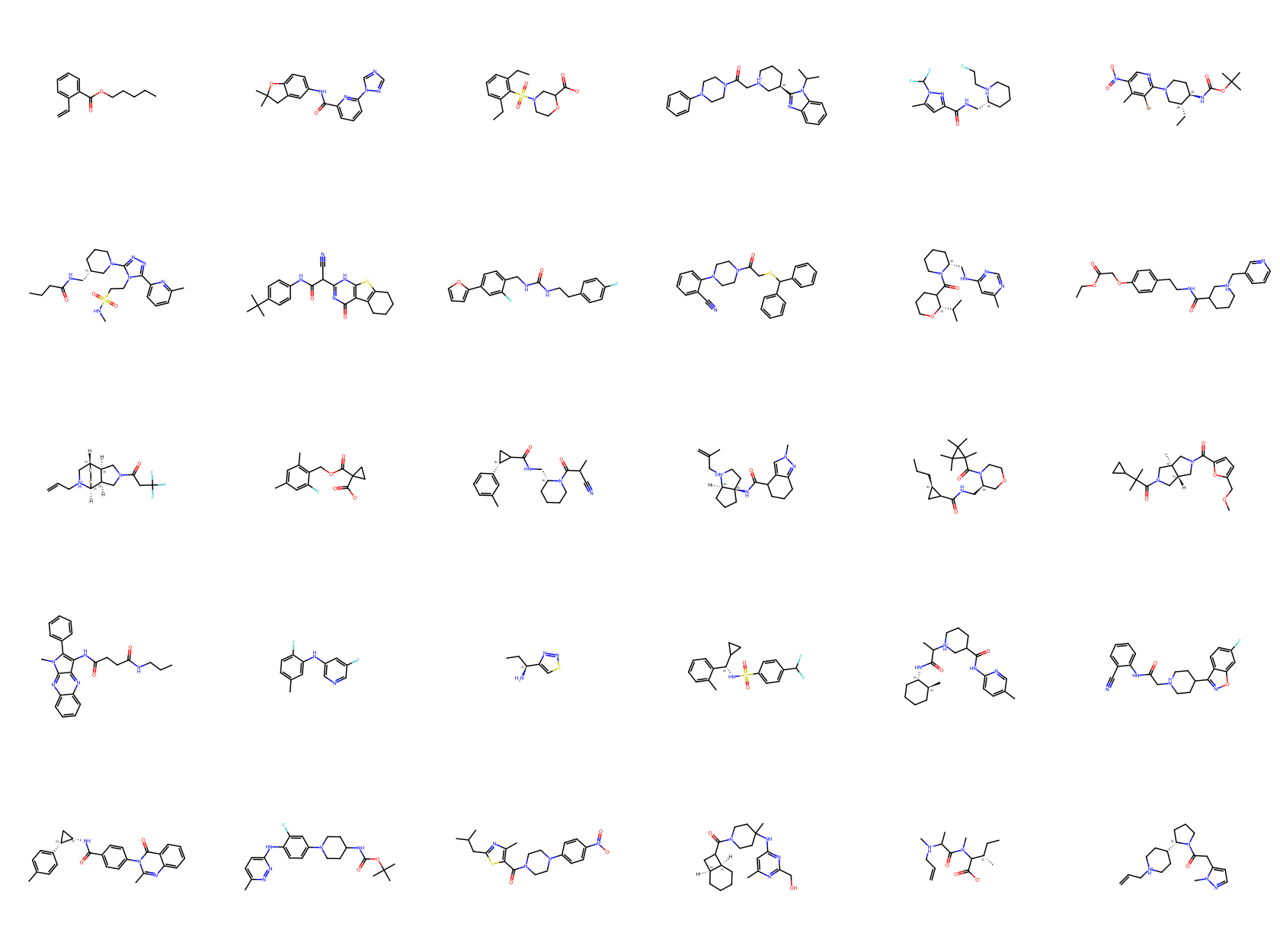}
    \caption{Randomly selected samples of \textit{de novo} generated molecules using SAFE.}
    \label{fig:sample-denovo}
\end{figure}

\begin{figure}[!htb]
    \centering
    \includegraphics[width=1\columnwidth]{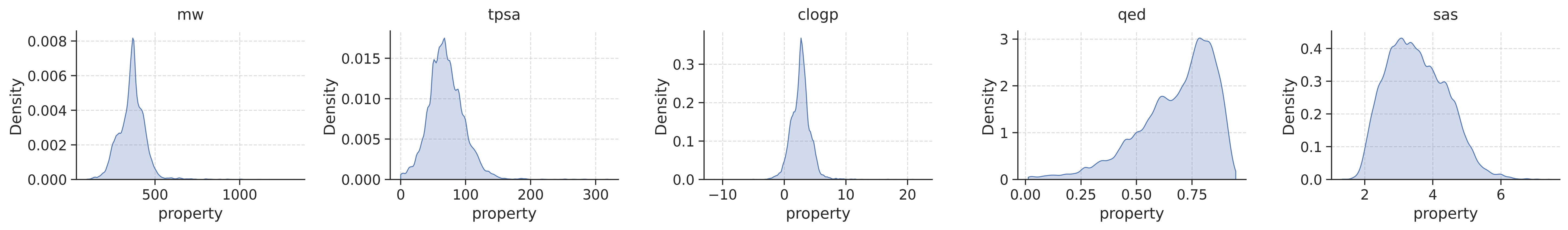}
    \caption{The molecular property distribution for 10,000 molecules generated with SAFE-GPT demonstrates that SAFE-GPT can generate molecules with diverse physicochemical properties, spanning beyond traditional drug-like molecules.}
    \label{fig:sample-prop-dist}
\end{figure}

\subsection{Comparison between SAFE and Group SELFIES}
\label{appendix:safe-groupselfies-comparison}

Both SAFE and Group SELFIES are molecular string representations capable of encoding fragments. In SAFE, fragments are denoted in groups of SMILES tokens separated by dots, while in Group SELFIES, fragments are tokens from a pre-defined grammar of chemical motifs (such as a token representing a toluene fragment). To compare their performance, we trained SAFE-GPT-20M and GSELFIES-GPT-20M on the MOSES dataset and evaluated them in \textit{de novo} molecule generation. We generated 10,000 molecules from each model and analyzed the distribution of molecular properties within these two sets to assess their efficacy.

\begin{figure}[!htb]
    \centering
    \includegraphics[width=1\columnwidth]{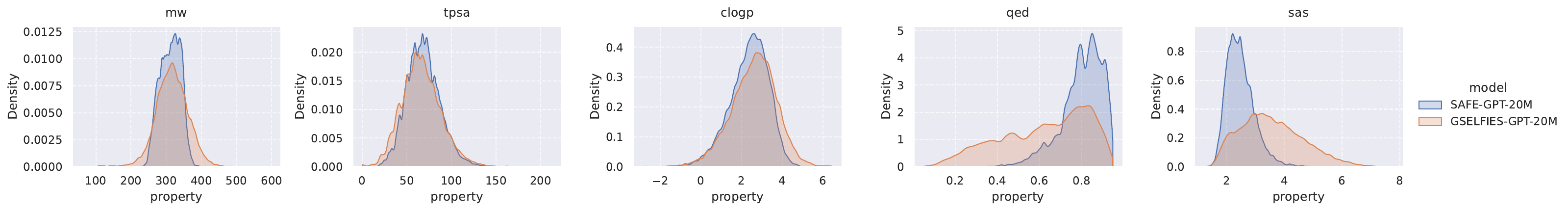}
    \caption{The molecular property distribution of molecules generated with SAFE-GPT-20M compared against molecules generated with GSELFIES-GPT-20M. }
    \label{fig:mol-prop-distribution-comparison}
\end{figure}

As seen in \autoref{fig:mol-prop-distribution-comparison}, molecules generated by SAFE-GPT-20M tend to exhibit higher QED (Quantitative Estimate of Drug-likeness) scores, indicating higher degree of drug-likeness, and lower SA (Synthetic Accessibility)  scores, indicating better synthetic feasibility. 

\begin{figure}[!htb]
    \centering
    \begin{subfigure}{0.51\columnwidth}
        \centering
    \includegraphics[width=\linewidth]
    {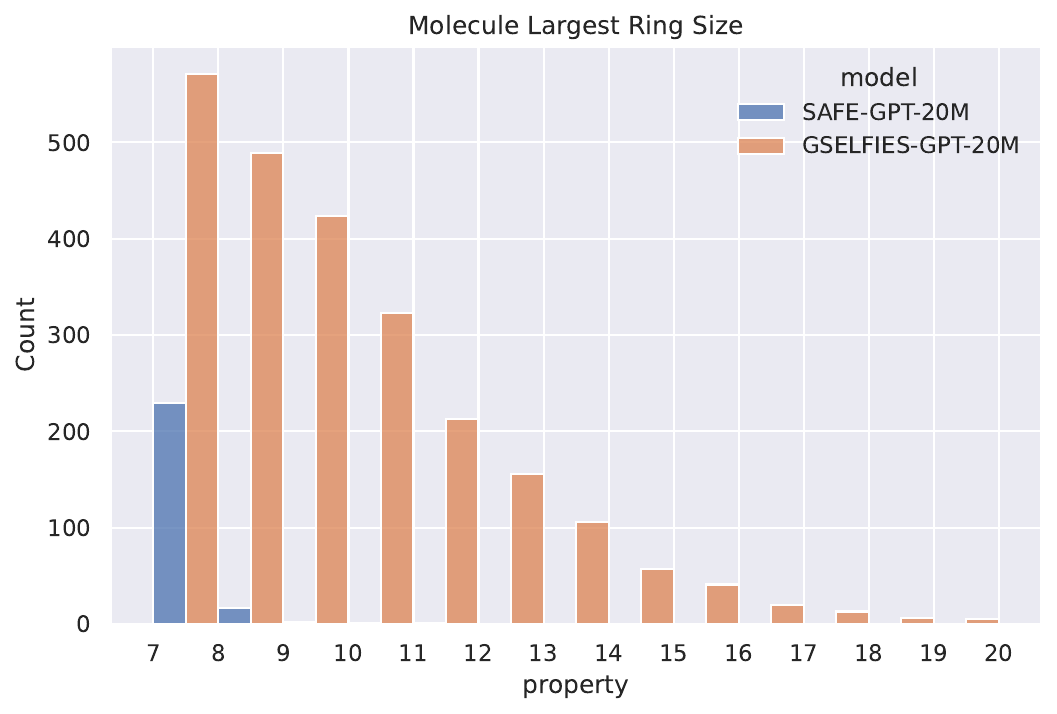}
    \caption{}
        \label{fig:max-ring-denovo-dist}
    \end{subfigure}
     \begin{subfigure}{0.47\columnwidth}
        \centering
    \includegraphics[width=\linewidth]{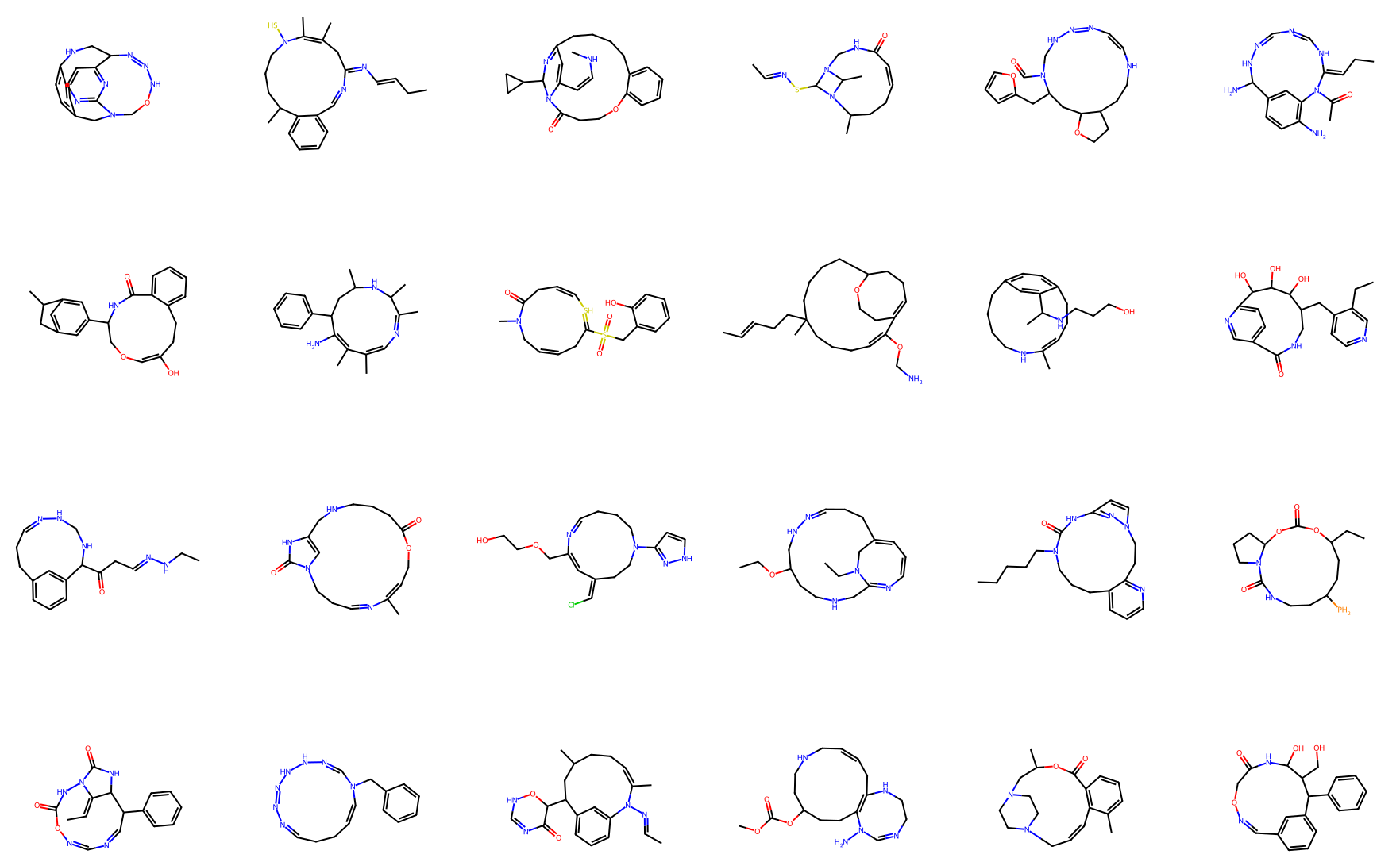}
    \caption{}
    \label{fig:max-ring-denovo-sample}
    \end{subfigure}%
    \caption{Distribution of the largest ring size (> 6 atoms) count in molecules generated with SAFE-GPT-20M compared against molecules generated with GSELFIES-GPT-20M. \textbf{(a)} GSELFIES-GPT-20M tends to generate molecules with ring sizes exceeding 8 atoms more frequently. \textbf{(b)} Examples of large ring molecules produced by GSELFIES-GPT-20M, illustrating their tendency towards non-druglike and chemically unstable structures.}
    \label{fig:max-ring-size-distribution}
\end{figure}

We further investigate the differences in the molecules generated by the two models by comparing the distributions of the largest ring size of each molecule. As shown on \autoref{fig:max-ring-size-distribution}, the model trained using the Group SELFIES notation frequently generate molecules with large and unstable ring structures.

We did not make further experiments and comparisons for the fragment-constrained generation tasks (such as linker design and scaffold decoration) as non-trivial adaptations would have to be made to the Group SELFIES notation, training process and molecular sampling, which could be explored in future works.

\subsection{Optimizing CNS penetration for EGFR inhibitors}
\label{cns-optimization-appendix}
Most existing small molecule treatments struggle to effectively penetrate the central nervous system (CNS) due to difficulties in breaching the blood-brain barrier (BBB). Notably, three well-known EGFR inhibitors (afatinib, gefitinib, and erlotinib), all sharing the same scaffold, exhibit generally low CNS penetration rates, with reported values respectively falling below 1\%, in the range of 1\%–3\%, and in the range of 3\%–6\%. The ability of a small molecule to penetrate the CNS is often associated with specific physicochemical properties such as CLogD, TPSA, and Molecular Weight. Various scoring systems have been developed to assess this ability. Notably, our findings indicate a correlation between the CNS MPO score~\citep{wager2016central} and the experimental penetration rates for these three EGFR inhibitors. \\

\begin{figure}[!htb]
\centering
\includegraphics[width=0.9\columnwidth]{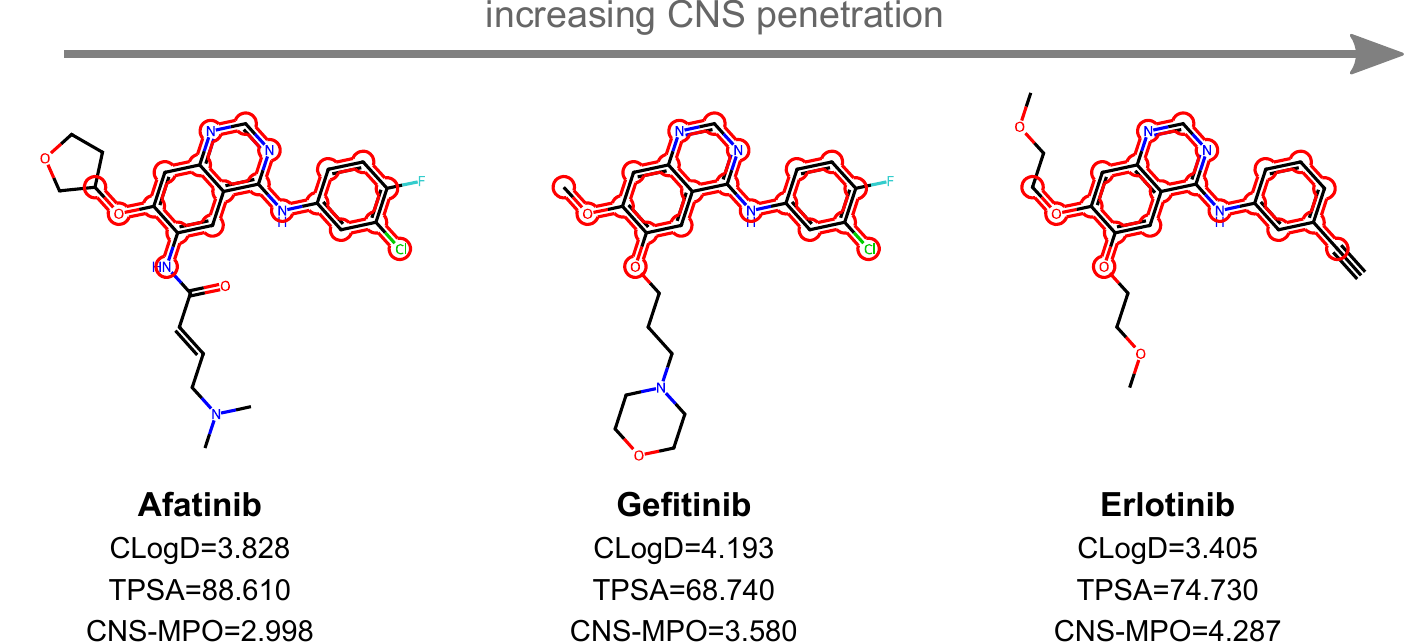}
	\caption{\small Existing EGFR inhibitors and their CNS profile}
	\label{fig:cns-penetration}
\end{figure}

\subsection{Fragment-constrained design results}
\label{appendix:frag-benchmark}

We uses a set of 10 drugs, including \textbf{Cyclothiazide}, \textbf{Maribavir}, \textbf{Spirapril}, \textbf{Baricitinib}, \textbf{Eliglustat}, \textbf{Erlotinib}, \textbf{Futibatinib}, \textbf{Lesinurad}, \textbf{Liothyronine}, and \textbf{Lovastatin}. These drugs were chosen as the basis for our fragment-constrained generative design tasks. From each drug, we extracted the main scaffold with attachment points, fragments that serve as side chains, a starting motif, and a core substructure.  These components were then respectively used as input for scaffold decoration, linker design / scaffold morphing,  motif extension, and superstructure generation, each with its specific objective. The details of the selected drugs and their corresponding inputs for each task can be found in \autoref{table:fragment-dataset}. It should be noted that linker design and scaffold morphing are two very similar tasks that share the same inputs. In our implementation, the only difference between them lies in the constraints imposed during sampling. For linker design, we employ a constrained beam search to ensure the presence of every fragment in the final molecules. In contrast, for scaffold morphing, new molecules are generated from each fragment with connectivity constraints, after which the scaffold is inferred and linked to the other fragments.

\begin{table}[!htb]
\small
\centering
\begin{threeparttable}
\caption{List of 10 known drugs and corresponding inputs used by SAFE-GPT for the fragment-constrained benchmark.\\}
\label{table:fragment-dataset}
\begin{tabular}{>{\centering\arraybackslash}m{2.1cm} >{\centering\arraybackslash}m{1.9cm} >{\centering\arraybackslash}m{1.9cm} >{\centering\arraybackslash}m{1.9cm} >{\centering\arraybackslash}m{1.9cm} >{\centering\arraybackslash}m{1.9cm}} 
\toprule
 \textbf{Name} & \textbf{Structure} & \textbf{Linker Design\tnote{*}} & \textbf{Scaffold Decoration}   & \textbf{Motif Extension}  & \textbf{Superstructure} \\ 
 \midrule
BARICITINIB
& \adjustimage{width=1.9cm}{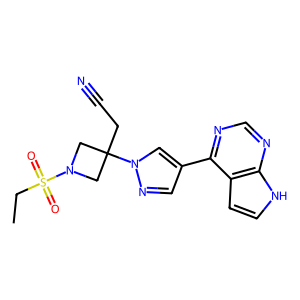}
& \adjustimage{width=1.9cm}{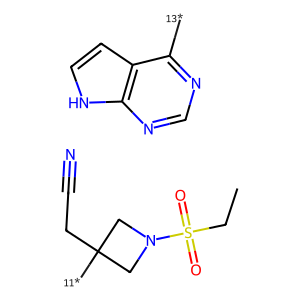}
& \adjustimage{width=1.9cm}{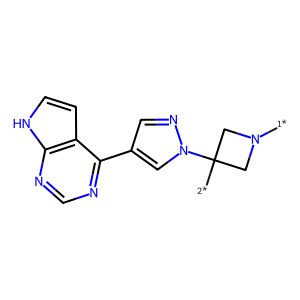}
& \adjustimage{width=1.9cm}{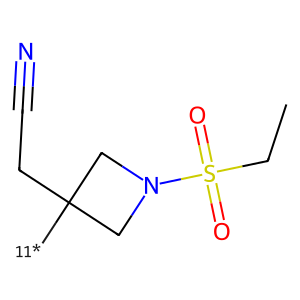}
& \adjustimage{width=1.9cm}{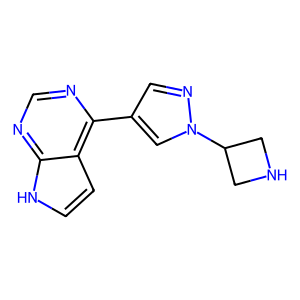} \\
CYCLOTHIAZIDE
& \adjustimage{width=1.9cm}{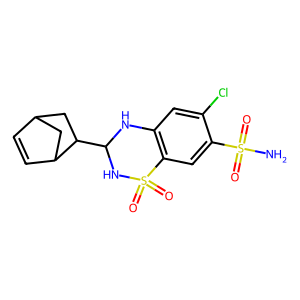}
& \adjustimage{width=1.9cm}{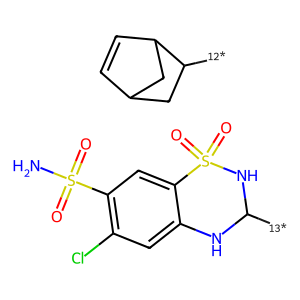}
& \adjustimage{width=1.9cm}{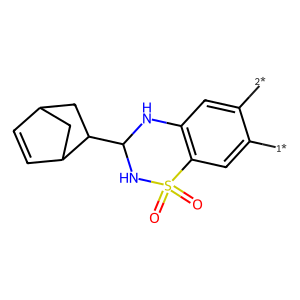}
& \adjustimage{width=1.9cm}{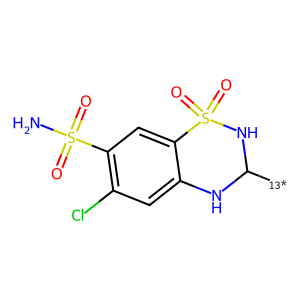}
& \adjustimage{width=1.9cm}{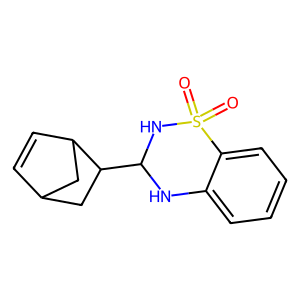} \\
ELIGLUSTAT
& \adjustimage{width=1.9cm}{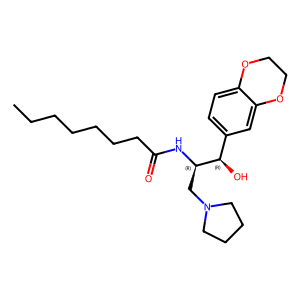}
& \adjustimage{width=1.9cm}{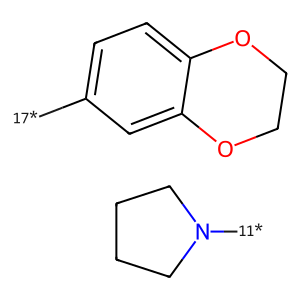}
& \adjustimage{width=1.9cm}{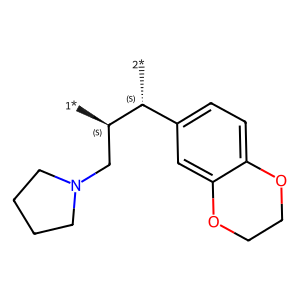}
& \adjustimage{width=1.9cm}{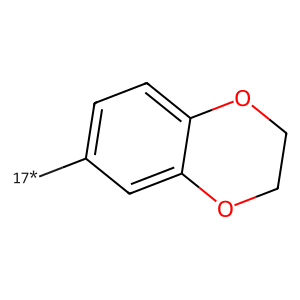}
& \adjustimage{width=1.9cm}{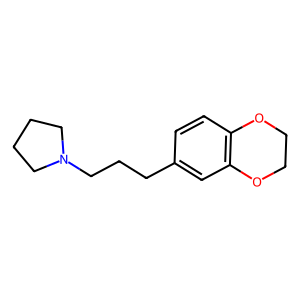} \\
ERLOTINIB
& \adjustimage{width=1.9cm}{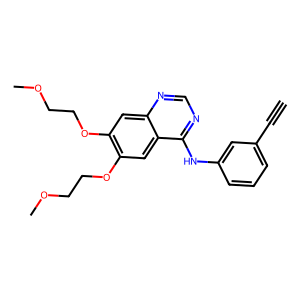}
& \adjustimage{width=1.9cm}{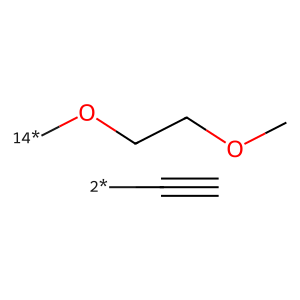}
& \adjustimage{width=1.9cm}{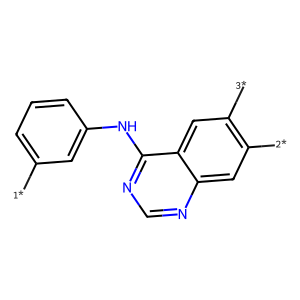}
& \adjustimage{width=1.9cm}{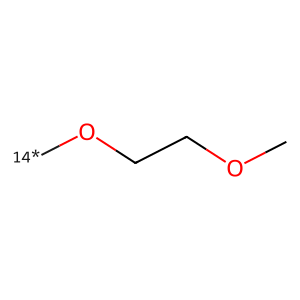}
& \adjustimage{width=1.9cm}{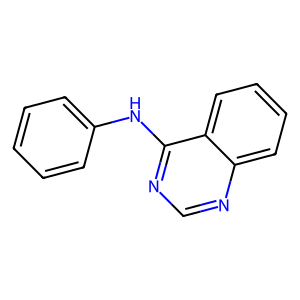} \\
FUTIBATINIB
& \adjustimage{width=1.9cm}{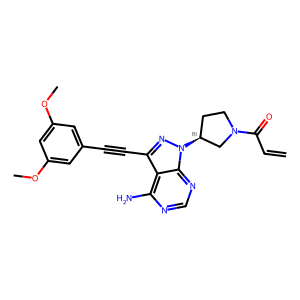}
& \adjustimage{width=1.9cm}{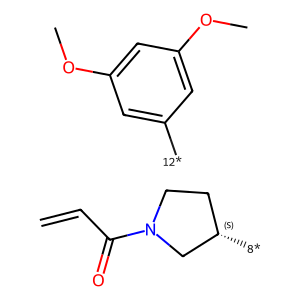}
& \adjustimage{width=1.9cm}{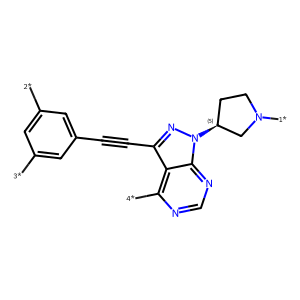}
& \adjustimage{width=1.9cm}{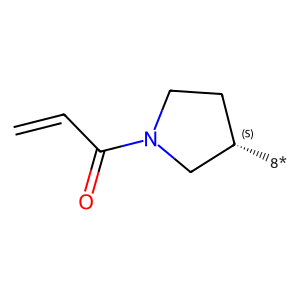}
& \adjustimage{width=1.9cm}{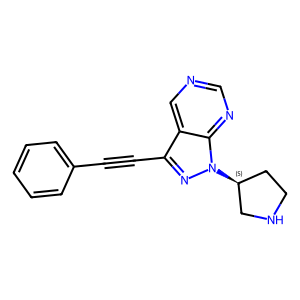} \\
LESINURAD
& \adjustimage{width=1.9cm}{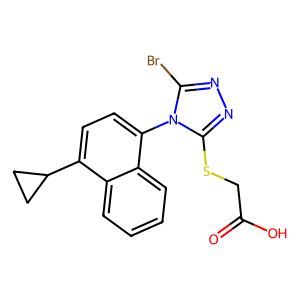}
& \adjustimage{width=1.9cm}{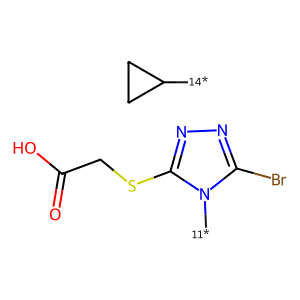}
& \adjustimage{width=1.9cm}{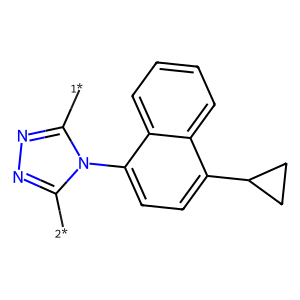}
& \adjustimage{width=1.9cm}{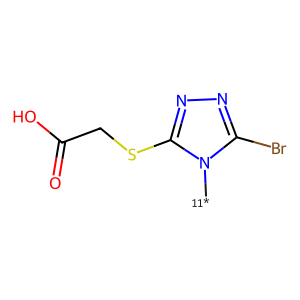}
& \adjustimage{width=1.9cm}{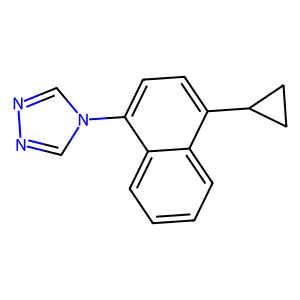} \\
LIOTHYRONINE
& \adjustimage{width=1.9cm}{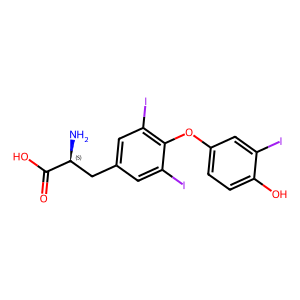}
& \adjustimage{width=1.9cm}{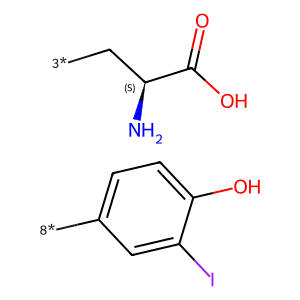}
& \adjustimage{width=1.9cm}{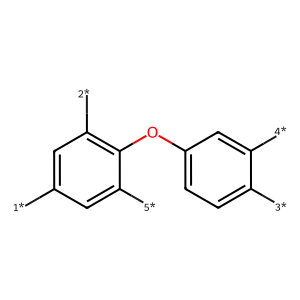}
& \adjustimage{width=1.9cm}{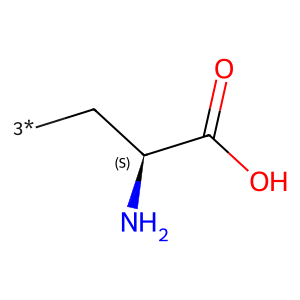}
& \adjustimage{width=1.9cm}{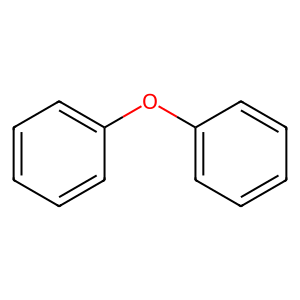} \\
LOVASTATIN
& \adjustimage{width=1.9cm}{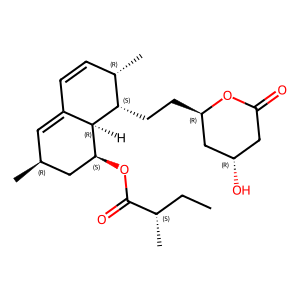}
& \adjustimage{width=1.9cm}{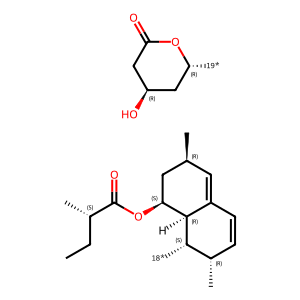}
& \adjustimage{width=1.9cm}{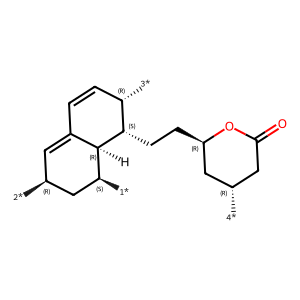}
& \adjustimage{width=1.9cm}{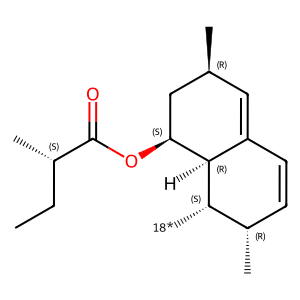}
& \adjustimage{width=1.9cm}{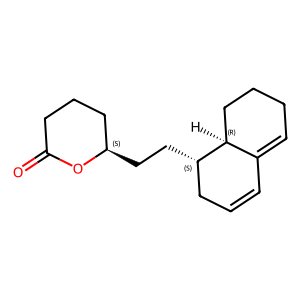} \\
MARIBAVIR
& \adjustimage{width=1.9cm}{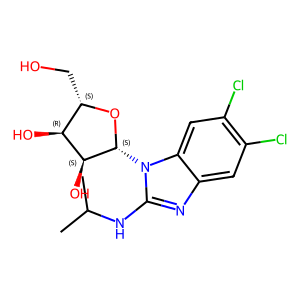}
& \adjustimage{width=1.9cm}{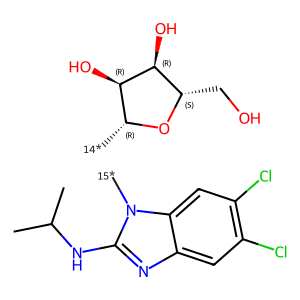}
& \adjustimage{width=1.9cm}{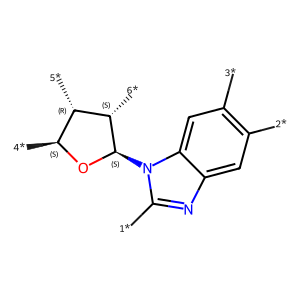}
& \adjustimage{width=1.9cm}{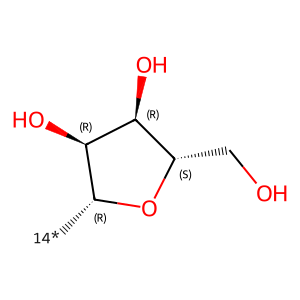}
& \adjustimage{width=1.9cm}{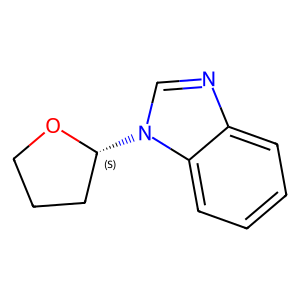} \\
SPIRAPRIL
& \adjustimage{width=1.9cm}{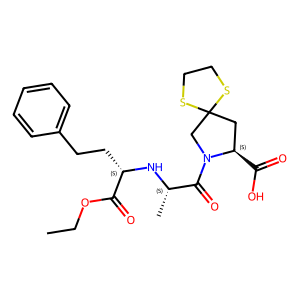}
& \adjustimage{width=1.9cm}{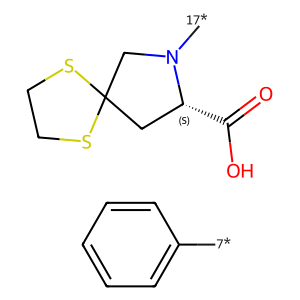}
& \adjustimage{width=1.9cm}{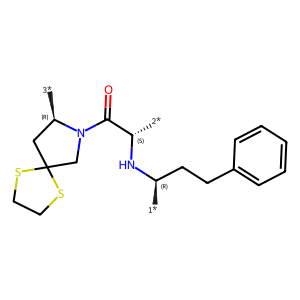}
& \adjustimage{width=1.9cm}{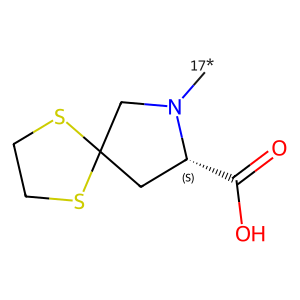}
& \adjustimage{width=1.9cm}{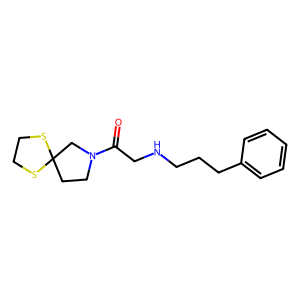} \\
\bottomrule
\end{tabular}%
\begin{tablenotes}
  \small
  \item[*] the linker design and scaffold morphing task share the same input fragments.
\end{tablenotes}
\end{threeparttable}
\end{table}

\begin{table}[!h]
\small
\centering
\caption{Examples of generated samples under fragment-constraints for the Maribavir structure\newline}
\label{table:example-sample-fragment-design}
\begin{tabular}{>{\arraybackslash}m{2cm} >{\arraybackslash}m{10cm}} 
\toprule
\textbf{Task} & \textbf{Generated samples}\\
 \midrule
Linker design& \includegraphics[width=0.75\columnwidth]{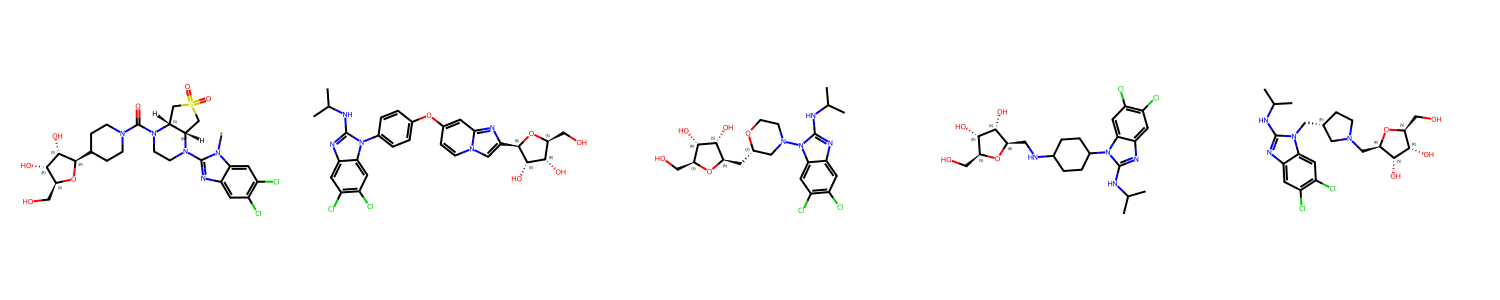} \\
Scaffold morphing & \includegraphics[width=0.75\columnwidth]{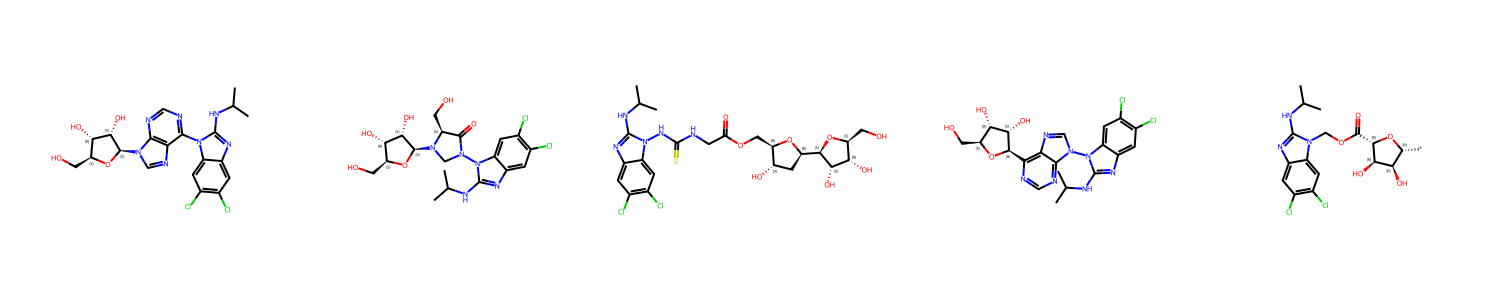} \\
Motif extension & \includegraphics[width=0.75\columnwidth]{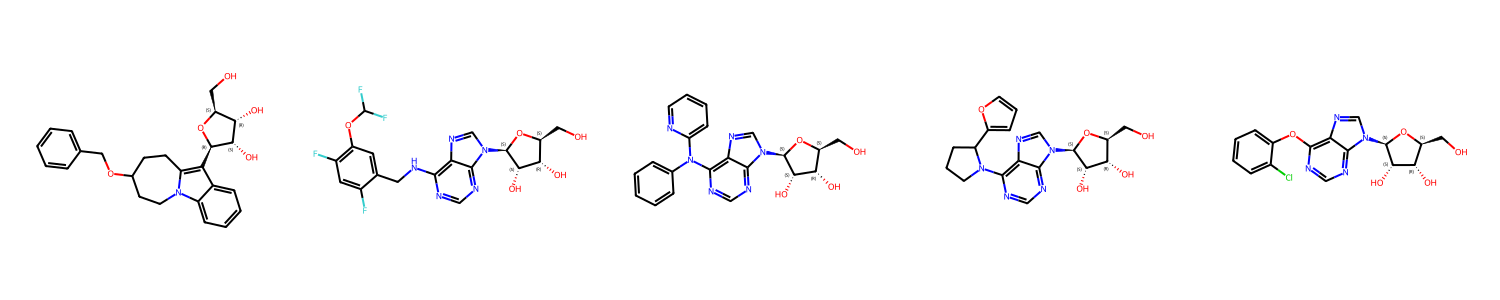} \\
Scaffold decoration& \includegraphics[width=0.75\columnwidth]{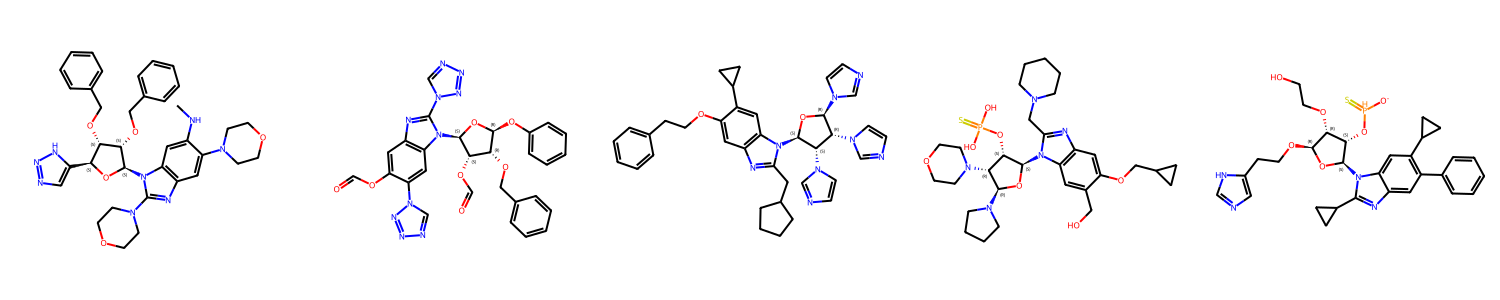} \\
Superstructure & \includegraphics[width=0.75\columnwidth]{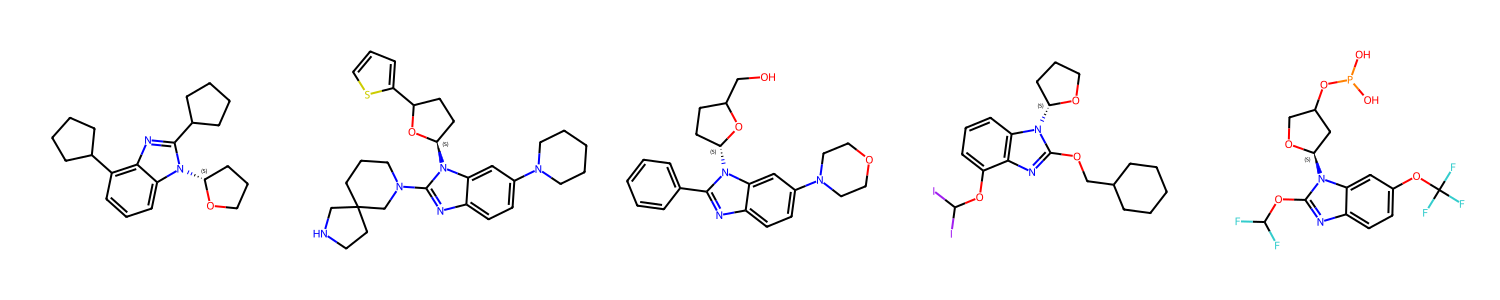} \\
 \bottomrule
\end{tabular}%
\end{table}




\end{document}